%% file: main.tex
\documentclass{article}

\usepackage{microtype}
\usepackage{graphicx}
\usepackage{subcaption}
\usepackage{booktabs} 
\usepackage{multirow} 
\usepackage{array} 
\usepackage[export]{adjustbox} 

\usepackage{hyperref}

\usepackage[preprint]{icml2026}

\usepackage{amsmath}
\usepackage{amssymb}
\usepackage{mathtools}
\usepackage{amsthm}

\usepackage{algorithm}
\usepackage{algorithmic}

\providecommand{\Return}{\textbf{return}}

\usepackage[capitalize,noabbrev]{cleveref}

\theoremstyle{plain}

\theoremstyle{definition}

\theoremstyle{remark}

\usepackage[textsize=tiny]{todonotes}

\newcommand{\imgpath}{fig/generated_images}
\newcommand{\imgw}{0.13\textwidth}

\icmltitlerunning{Look-Ahead and Look-Back Flows: Training-Free Image Generation with Trajectory Smoothing}

\begin{document}

\twocolumn[
  \icmltitle{Look-Ahead and Look-Back Flows: Training-Free Image Generation with Trajectory Smoothing}



  \icmlsetsymbol{equal}{*}

  \begin{icmlauthorlist}
    \icmlauthor{Yan Luo}{hv}
    \icmlauthor{Henry Huang}{hv}
    \icmlauthor{Todd Y. Zhou}{hv}
    \icmlauthor{Mengyu Wang}{hv}
  \end{icmlauthorlist}

  \icmlaffiliation{hv}{Harvard AI and Robotics Lab, Harvard University}

  \icmlcorrespondingauthor{Mengyu Wang}{mengyu\_wang@meei.harvard.edu}

  \icmlkeywords{Flow Matching, Training-Free Method, Text-to-Image Generation, Trajectory Smoothing}

  \vskip 0.3in
]



\printAffiliationsAndNotice{}  

\input{sec/abs}

\input{sec/inro}
\input{sec/related}

\input{sec/method}

\input{sec/expr}

\input{sec/conclusion}

\section*{Impact Statement}

This work introduces Look-Ahead and Look-Back, training-free mechanisms that enhance the numerical stability and trajectory fidelity of flow-based generative models. By correcting discretization errors such as divergence and overshooting, these methods ensure that the generated output adheres more strictly to the semantic intent and distribution defined by the pre-trained velocity field.

There are many potential societal consequences of our work, none of which we feel must be specifically highlighted here. Because our contribution is restricted to a numerical stabilization technique that does not modify model parameters or training data, it introduces no new ethical dimensions or risks independent of the pre-trained model weights.

\bibliography{main}
\bibliographystyle{icml2026}

\newpage
\appendix
\onecolumn
\input{sec/appendix}



\end{document}

%% file: sec/abs.tex
\begin{abstract}
Recent advances have reformulated diffusion models as deterministic ordinary differential equations (ODEs) through the framework of flow matching, providing a unified formulation for the noise-to-data generative process. Various training-free flow matching approaches have been developed to improve image generation through flow velocity field adjustment, eliminating the need for costly retraining. However, Modifying the velocity field $v$ introduces errors that propagate through the full generation path, whereas adjustments to the latent trajectory $z$ are naturally corrected by the pretrained velocity network, reducing error accumulation. In this paper, we propose two complementary training-free latent-trajectory adjustment approaches based on future and past velocity $v$ and latent trajectory $z$ information that refine the generative path directly in latent space. We propose two training-free trajectory smoothing schemes: \emph{Look-Ahead}, which averages the current and next-step latents using a curvature-gated weight, and \emph{Look-Back}, which smoothes latents using an exponential moving average with decay. We demonstrate through extensive experiments and comprehensive evaluation metrics that the proposed training-free trajectory smoothing models substantially outperform various state-of-the-art models across multiple datasets including COCO17, CUB-200, and Flickr30K.

\end{abstract}

%% file: sec/inro.tex
\section{Introduction}

\begin{figure}[t!]
  \centering
  \includegraphics[width=0.3\textwidth]{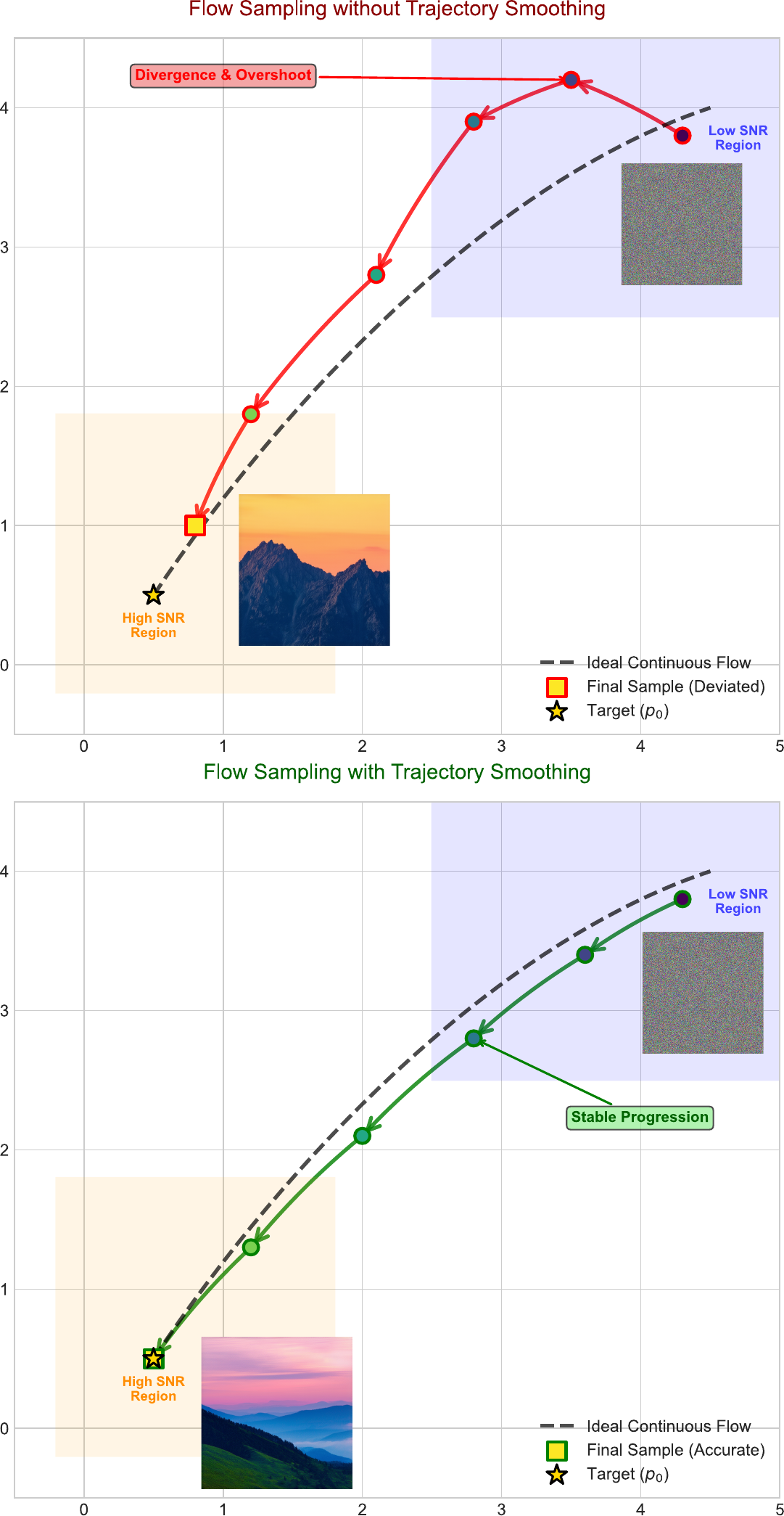}
  \caption{Conceptual illustration of training-free trajectory smoothing for flow sampling. Without trajectory smoothing (top), backward integration of the flow ODE suffers divergence and overshoot in low Signal-to-Noise Ratio (SNR) regions, causing the discrete trajectory to deviate from the ideal continuous flow and producing final samples that inaccurately reach the target distribution. With the trajectory smoothing mechanism (bottom), the trajectory maintains robust fidelity to the ideal continuous flow across both low and high SNR regions, ensuring stable progression and accurate convergence.
  }
  \label{fig:two-subfigs}
\end{figure}


Recently, diffusion models have been reformulated within the mathematical framework of ordinary differential equations (ODEs), known as \emph{flow matching}, which defines a deterministic ODE-governed formulation of noise-to-data generative process~\cite{lipman2023flow,chen2024flow,albergo2023stochastic, luo2025curveflow}. Among these ODE-based approaches, \emph{Rectified Flow} has emerged as a particularly effective formulation that uses a linear and constant-velocity transport between noise and data as a training target~\cite{liu2022rectified,liu2023flow,esser2024scaling}. This simplification not only leads to more stable training and faster convergence~\cite{lee2024improving,wang2024straightness} but also enables efficient few-step generation and straightforward distillation of pretrained flow models~\cite{kornilov2024optimal,liu2024instaflow}. More recently, a series of \emph{training-free} approaches have further advanced this line of work by adapting pretrained flow models through velocity re-parameterization or trajectory refinement, eliminating the need for costly retraining~\cite{wang2025training,bu2025hiflow,li2025self,jin2025flops}. These methods are especially appealing because they achieve substantial improvements in image generation fidelity and image editing capability with minimal computational overhead~\cite{wang2025taming,kulikov2025flowedit,avrahami2025stableflow}, making high-quality generation and editing accessible even under limited resources.

Most training-free flow methods focus on image editing using pretrained rectified-flow or diffusion models without retraining. Stable Flow~\cite{avrahami2025stableflow} identifies vital layers for improved editing; RF-Edit~\cite{wang2024taming} preserves structure by reusing stored self-attention features; RF-Inversion~\cite{rout2024semantic} uses an optimal-control adjustment of the image-to-noise trajectory for faithful zero-shot editing. FlowEdit~\cite{kulikov2025flowedit} constructs a direct flow between source- and target-prompt images for inversion-free editing. FlowChef~\cite{patel2025flowchef} nudges the latent trajectory using a task loss on the predicted clean image. SplitFlow~\cite{yoon2025splitflow} decomposes a target prompt into semantic subflows and re-assembles them for higher-fidelity, inversion-free editing.

By contrast, only a few training-free flow methods focus on improving general image generation. Rectified Diffusion~\cite{wang2024rectified} enhances fidelity by finetuning on synthetic samples. HiFlow~\cite{bu2025hiflow} builds a virtual high-resolution reference flow to guide generation, while OC-Flow~\cite{wang2025training} steers trajectories via reward-driven control velocities. A-Euler~\cite{jin2025flops} enforces near-linear velocity by decomposing it into drift and residual components, and Self-Guidance~\cite{li2025self} stabilizes denoising by smoothing velocity with past-step corrections.

In this paper, we propose two complementary training-free latent-trajectory adjustment approaches based on future and past velocity $v$ and latent trajectory $z$ information that refine the generative path directly in latent space, in contrast to prior methods that modify the velocity field. Modifying the velocity field $v$ introduces errors that propagate along the entire generation path, whereas adjustments to the latent trajectory $z$ are subsequently regularized by the pretrained velocity network, limiting error accumulation.


The contributions of our paper are as follows:

\begin{itemize}

\item We propose a training-free \emph{Look-Ahead} scheme that smoothes the latent trajectory with weighted average of current $z$ and next-step $z$ gated by spatial curvature. 


\item We introduce a training-free \emph{Look-Back} scheme that smoothes the latent trajectory with exponential moving average of latent state $z$ with a decay.



\item We demonstrate through extensive experiments and comprehensive evaluation metrics that the proposed training-free trajectory smoothing models substantially outperforms various state-of-the-art models.

\end{itemize}

%% file: sec/related.tex
\section{Related Work}


\paragraph{Flow Matching and Its Variants}

Flow Matching~\cite{lipman2023flow} formulates generative modeling as learning a continuous-time velocity field that transports samples from a simple prior distribution to the target data distribution via an ordinary differential equation (ODE). 
Subsequent works have extended this framework to various geometric and architectural settings, such as Flow Matching on general geometries~\cite{chen2024flow}, discrete domains~\cite{gat2024discrete}, and rigid-body manifolds through SE(3)-Stochastic Flow Matching~\cite{bose2024se3}. The Rectified Flow formulation~\cite{liu2023flow} further simplifies the probability path by enforcing a linear and constant-velocity transport between data and noise, leading to faster convergence and trajectory-straightened flows. Follow-up studies explore optimal transport straightness~\cite{wang2024straightness}, one-step trajectory learning~\cite{kornilov2024optimal}, and improved training objectives~\cite{lee2024improving}, while others incorporate adaptive control~\cite{wang2025training}, reinforcement learning~\cite{liu2025flow}, or velocity decomposition~\cite{jin2025flops} to enhance stability and efficiency. Training-free rectified-flow variants, such as HiFlow~\cite{bu2025hiflow} and Self-Guidance~\cite{li2025self}, demonstrate that pretrained flow models can be refined through flow-aligned or self-consistent guidance without retraining.
Collectively, these advances establish Rectified Flow as a unifying paradigm for deterministic generation, underscoring the widespread adoption of Stable Diffusion v3.5~\cite{esser2024scaling} in modern generative modeling pipelines.

\paragraph{Training-Free Flow-Based Image Generation Methods}
Most existing training-free flow models primarily target image editing \cite{avrahami2025stableflow,wang2024taming,rout2024semantic,kulikov2025flowedit,patel2025flowchef,yoon2025splitflow}, where pretrained flow matching models are adapted without additional training to enable high-quality and controllable visual modifications.
By contrast, relatively few training-free flow-based methods target improvements in general image generation. For instance, Rectified Diffusion~\cite{wang2024rectified} Rectified Diffusion demonstrates that leveraging synthetic samples generated by a pretrained diffusion or flow-matching model to perform a secondary finetuning stage leads to consistent improvements in downstream generation fidelity and sample quality. HiFlow~\cite{bu2025hiflow} establishes a virtual reference flow within the high-resolution space that effectively captures the characteristics of low-resolution flow information, offering guidance for high-resolution generation. OC-Flow~\cite{wang2025training} maximizes a chosen image-level reward such as CLIP similarity and aesthetic score to optimally steer the flow trajectory by adding a term of control velocity to produce images that satisfy the desired condition. Training-free velocity-modification methods such as OC-Flow show strong potential for improving image generation quality by directly steering the flow trajectory without retraining. Along this line of research, A-Euler~\cite{jin2025flops} accelerates few-step sampling by adaptively decomposing the velocity field into a linear drift and a temporally-suppressed residual, effectively enforcing a near-linear velocity trajectory. Self-Guidance~\cite{li2025self} smooths the sampling trajectory by correcting the current velocity using a weighted difference from the noisier immediate-past velocity, leading to more stable denoising and improved image generation quality. Both A-Euler and Self-Guidance temporally smooth the velocity field to stabilize the denoising trajectory and improve image generation quality.

%% file: sec/method.tex
\section{Problem Statement: Training-Free Trajectory Smoothing}
\label{sec:problem_statement}

Let $p_0$ denote the target data distribution and $p_1=\mathcal{N}(0,I)$ the noise prior.
Flow-based generative models, including rectified flow and its variants, define a continuous probability path $(p_t)_{t\in[0,1]}$ governed by the ordinary differential equation (ODE)
\begin{equation}
\frac{dz_t}{dt} = v_\Theta(z_t,t,c), 
\qquad z_1 \!\sim\! p_1,~z_0\!\sim\! p_0,
\label{eq:flow_ode_problem}
\end{equation}
where $v_\Theta$ denotes a learned velocity field parameterized by $\Theta$ and conditioned on optional context $c$.

During inference, the path is integrated backward in time using discrete timesteps $\{t_k\}_{k=0}^{K}$. 
However, the numerical integration of this flow ODE is often unstable due to high local curvature in the learned velocity field $v_\Theta$, stiffness and oscillations in regions of low signal-to-noise ratio (SNR), and the limited step sizes prescribed by schedulers originally tuned for diffusion models. 
Such instabilities manifest as divergence, overshooting, or loss of trajectory fidelity, leading to degraded sample quality and requiring either retraining or costly higher-order solvers.

\paragraph{Objective}
The goal is to design a training-free, lightweight stabilization mechanism that improves numerical stability and trajectory fidelity during backward integration of the flow ODE, operates without additional model evaluations or retraining of $v_\Theta$, and remains fully compatible with native scheduler configurations used in diffusion or rectified-flow sampling. 
This mechanism should enhance robustness while preserving the efficiency and generality of standard inference pipelines.

\paragraph{Formal Optimization View}
Given a discrete trajectory $\{z_k\}$ governed by the update rule
\begin{equation}
z_{k+1} = \Phi_\Theta(z_k, t_k, \eta_k, c),
\label{eq:discrete_update}
\end{equation}
we seek a correction operator $\mathcal{R}$ such that
\begin{align}
\begin{split}
&\min_{\mathcal{R}}~ 
\mathbb{E}_{t,z}\!\left[\|z_{k+1}^{\mathcal{R}} - \psi_t(z_1)\|_2^2 \right] \\
&\text{s.t.} \quad 
z_{k+1}^{\mathcal{R}} = \mathcal{R}\bigl(\Phi_\Theta(z_k,t_k,\eta_k,c)\bigr),
\end{split}
\label{eq:stabilization_opt}
\end{align}
where $\psi_t(\cdot)$ denotes the ideal continuous flow trajectory.
To achieve stable and faithful integration without retraining, we introduce two training-free correction schemes for $\mathcal{R}$.
The first, the \emph{Look-Ahead} scheme, stabilizes the trajectory by referencing future curvature trends to anticipate deviations and adjust the step accordingly. 
The second, the \emph{Look-Back} scheme, achieves complementary stability by referencing past averaged states, effectively damping high-frequency oscillations while maintaining trajectory fidelity. 
Both mechanisms adaptively regulate the integration process to preserve the expected flow dynamics of the learned velocity field $v_\Theta$.

\begin{figure*}[t!]
  \centering
  \includegraphics[width=1\textwidth]{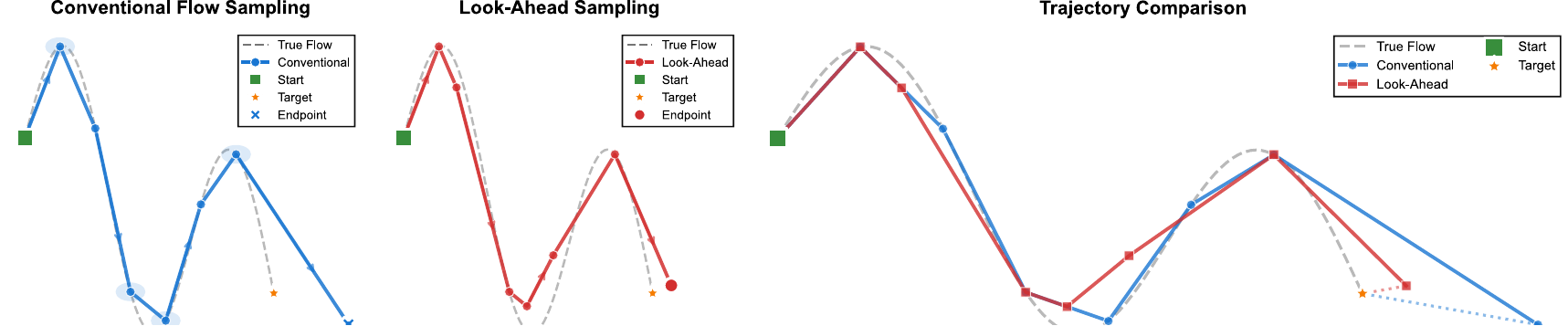}
  \caption{Schematic view of the proposed look-ahead sampling. Conventional flow sampling always takes full steps, which can overshoot in regions of high curvature and lead to a large deviation from the target. In contrast, the Look-Ahead scheme adaptively interpolates based on local curvature, modulating step sizes to better follow the underlying flow and achieve a significantly smaller endpoint error.}
  \label{fig:lookahead}
\end{figure*}

\begin{figure}[t!]
  \centering
  \includegraphics[width=0.5\textwidth]{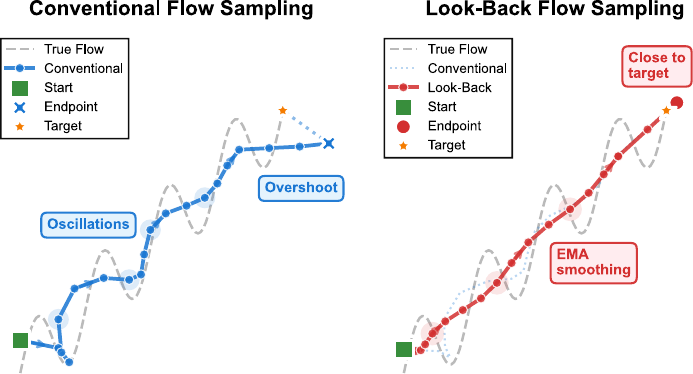}
  \caption{Schematic view of the proposed look-back sampling. Conventional sampling exhibits oscillations and overshoots the target, while Look-Back produces a smooth trajectory through exponential state averaging.}
  \label{fig:lookback}
\end{figure}

\section{Look-Ahead Scheme for Flow Sampling}
\label{sec:pc_lookahead_implementation}

Let $p_0$ be the data distribution and $p_1=\mathcal{N}(0,I)$ the noise prior.
A rectified-flow model defines a probability path $(p_t)_{t\in[0,1]}$ governed by
\begin{equation}
\frac{dz_t}{dt}=v_\Theta(z_t,t,c),\qquad z_1\!\sim p_1,\;z_0\!\sim p_0,
\label{eq:flow_ode_imp}
\end{equation}
where $v_\Theta$ denotes a learned velocity field conditioned on context $c$.
Sampling integrates \eqref{eq:flow_ode_imp} backward on a grid
$1=t_0>\cdots>t_K=0$ with steps $\Delta_k=t_k-t_{k+1}>0$.
The forward Euler update is
\begin{equation}
z_{k+1}^{\text{Eul}}=z_k-\eta_k\,v_\Theta(z_k,t_k,c),
\label{eq:euler_imp}
\end{equation}
where $\eta_k$ follows the scheduler’s native step configuration.

To improve stability without retraining, we introduce a
\emph{training-free look-ahead} scheme that inspects the local velocity trend:
it accepts long steps in nearly straight directions but interpolates
conservatively under high curvature, keeping constant cost per step.

\paragraph{Predictor and Efficient Velocity Estimation}
At $(z_k,t_k)$, we evaluate the instantaneous velocity
$v_k=v_\Theta(z_k,t_k,c)$
and perform the scheduler’s native predictor step,
$\tilde z=\text{SchedulerStep}(z_k,v_k,t_k)$ with $\tilde t=t_k-\Delta_k$.
Instead of re-evaluating $v_\Theta(\tilde z,\tilde t,c)$,
we estimate the peek velocity by finite difference:
\begin{equation}
\tilde v\approx\frac{z_k-\tilde z}{\Delta_k}.
\label{eq:peek_estimated_imp}
\end{equation}
This captures the effective directional change implied by the scheduler
without doubling model evaluations.

\paragraph{Curvature-Gated Interpolation}
We quantify local curvature as the normalized directional deviation:
\begin{equation}
\kappa_k=\frac{\|\tilde v-v_k\|_2}{\|\tilde z-z_k\|_2+\varepsilon},
\label{eq:kappa_imp}
\end{equation}
where $\varepsilon=10^{-8}$ ensures numerical stability by preventing division-by-zero.
If $\kappa_k\!\le\!\tau_{\mathrm{curv}}$, the full step is accepted,
$z_{k+1}=\tilde z,\,t_{k+1}=\tilde t$.
Otherwise, a partial interpolation mitigates instability:
\begin{equation}
z_{k+1}=z_k+\gamma(\tilde z-z_k),\qquad t_{k+1}=\tilde t,
\end{equation}
where $\gamma\!\in\!(0,1)$ controls conservativeness.
The timestep always advances to preserve scheduler synchronization.
This curvature-gated interpolation adaptively balances efficiency and stability: when the local trajectory is smooth ($\kappa_k$ small), it takes confident, full steps to accelerate sampling; when curvature increases, the interpolation automatically dampens the update, preventing divergence or oscillation.
In essence, the gate acts as a lightweight, geometry-aware step controller that preserves trajectory fidelity without extra model evaluations.

\paragraph{Algorithm}
Algorithm \ref{alg:pc_lookahead_imp} summarizes the full inference routine.
It iteratively integrates the flow ODE using a scheduler-consistent predictor, estimates future velocity via a single finite-difference peek, and adaptively accepts or interpolates each step based on local curvature, achieving stability and efficiency without additional model evaluations, as illustrated in Figure \ref{fig:lookahead}.
\begin{algorithm}[H]
\caption{Look-Ahead Sampling with Curvature Gate}
\label{alg:pc_lookahead_imp}
\begin{algorithmic}[1]
\REQUIRE trained field $v_\Theta(\cdot,\cdot,c)$, scheduler timesteps $\{t_k\}_{k=0}^{K}$ with $\Delta_k=t_k-t_{k+1}$,
initial latent $z_0 \sim p_1$ at $t_0=1$,
curvature threshold $\tau_{\mathrm{curv}}>0$,
interpolation factor $\gamma\in(0,1)$,
numerical epsilon $\varepsilon$
\STATE $z\!\leftarrow\!z_0$, $t\!\leftarrow\!t_0$
\FOR{$k=0,1,\dots,K-1$}
    \STATE $v\!\leftarrow\!v_\Theta(z,t,c)$
    \STATE $\tilde z\!\leftarrow\!\text{SchedulerStep}(z,v,t)$,\;
           $\tilde t\!\leftarrow\!t-\Delta_k$
    \STATE $\tilde v\!\leftarrow\!(z-\tilde z)/\Delta_k$
    \STATE $\kappa\!\leftarrow\!\|\tilde v-v\|_2/(\|\tilde z-z\|_2+\varepsilon)$
    \IF{$\kappa \le \tau_{\text{curv}}$}
        \STATE $z\!\leftarrow\!\tilde z$ \COMMENT{accept full step}
    \ELSE
        \STATE $z\!\leftarrow\!z+\gamma(\tilde z-z)$ \COMMENT{interpolate}
    \ENDIF
    \STATE $t\!\leftarrow\!\tilde t$
\ENDFOR
\STATE \textbf{return} $z$
\end{algorithmic}
\end{algorithm}

\paragraph{Reduction to Conventional Flow Sampling}
When $\tau_{\mathrm{curv}}=\infty$ or $\gamma=1$, the curvature gate never triggers interpolation, it becomes identical to conventional flow sampling. 

\section{Look-Back Scheme for Flow Sampling}
\label{sec:lookback}

While the Look-Ahead scheme stabilizes integration by probing future trends,
the \emph{Look-Back} scheme complements it by peeking into the past.
It leverages exponentially averaged latents to damp high-frequency oscillations,
providing a memory-efficient and training-free stabilization mechanism.

\paragraph{Exponential State Averaging}
We maintain an exponentially weighted history of latent states:
\begin{equation}
\bar z_k = \gamma(t_k)\,\bar z_{k-1} + \bigl(1-\gamma(t_k)\bigr)\,z_k,
\quad \bar z_{-1}=z_0,
\label{eq:ema_state}
\end{equation}
where $\gamma(t)\!\in\![0,1)$ controls the decay rate.
The averaged state $\bar z_k$ acts as a low-variance memory of the trajectory,
filtering transient noise while preserving the coarse structure of the flow dynamics.
As $\gamma(t)$ decreases, $\bar z_k$ becomes more responsive to the current latent,
ensuring the averaging influence vanishes near convergence.
Exponential state averaging functions as a temporal low-pass filter, suppressing jitter during mid-range timesteps with high stochastic variance. This yields smoother latent paths and improved numerical robustness for rectified-flow dynamics.

\paragraph{Backward-Referenced yet Forward-Stabilizing}
Unlike the look-ahead method that extrapolates into the future,
the Look-Back scheme stabilizes integration by referencing past states.
Velocity is evaluated at a blended latent that combines the current state
with its exponentially smoothed history:
\begin{equation}
z_k^{\text{peek}}=(1-\lambda)\,z_k+\lambda\,\bar z_{k-1},
\qquad \lambda\!\in\![0,1].
\label{eq:peek}
\end{equation}
Here, $\bar z_{k-1}$ captures the slow manifold of recent dynamics,
serving as a denoised, low-frequency reference.
Although the mechanism peeks backward, its effect is
\emph{forward-stabilizing}: the blended latent aligns velocity evaluation
with the stable trajectory toward which the flow is converging.
The corresponding update reads
\begin{equation}
z_{k+1}=z_k-\eta_k\,v_\Theta(z_k^{\text{peek}},t_k,c),
\label{eq:update}
\end{equation}
where $v_\Theta$ is the learned velocity field.
This implicit damping mitigates overshooting and oscillatory motion without biasing the underlying flow manifold.
By peeking into the past, the sampler anticipates the stable manifold of the trajectory, reducing variance and promoting smooth convergence in stiff or noisy regions of the flow.

\paragraph{SNR-Aware Decay Scheduling}
To adapt the smoothing strength across noise levels,
we define $\gamma(t)$ as a logistic function of the log-SNR:
\begin{equation}
\gamma(t)=\gamma_{\max}\cdot\sigma\!\Big(\beta[\xi(t)-\xi_\ast]\Big),
\label{eq:gamma}
\end{equation}
where $\sigma$ denotes the sigmoid, $\beta$ controls transition steepness,
and $\xi(t)=\log(a_t^2/b_t^2)$ is the log-SNR derived from the generative process
$z_t=a_t x_0+b_t\epsilon$ with $\epsilon\!\sim\!\mathcal{N}(0,I)$.
For rectified flows, $a_t=1-t$ and $b_t=t$ yield
$\xi(t)=2\log\!\frac{1-t}{t}$;
for diffusion models, $a_t=\sqrt{\bar\alpha_t}$ and $b_t=\sqrt{1-\bar\alpha_t}$,
so $\xi(t)=\log(\bar\alpha_t/(1-\bar\alpha_t))$.
The midpoint $\xi_\ast$ marks maximal uncertainty:
when $\xi(t)\!<\!\xi_\ast$, $\gamma(t)$ stays high for strong smoothing;
as $\xi(t)\!\to\!\infty$, $\gamma(t)\!\to\!0$, recovering the native solver near convergence.

\paragraph{Algorithm}
Algorithm \ref{alg:lookback} outlines the complete inference procedure for the Look-Back sampling, as illustrated in Figure \ref{fig:lookback}.
At each step, it maintains an exponentially averaged latent state, peeks into this smoothed history to evaluate a stabilized velocity, and updates the current latent accordingly, achieving noise-averaged, training-free flow sampling with improved stability and smoothness.
\begin{algorithm}[H]
\caption{Look-Back Sampling for Flow Models}
\label{alg:lookback}
\begin{algorithmic}[1]
\REQUIRE $v_\Theta$, schedule $\{(t_k,\eta_k)\}_{k=0}^{K-1}$, decay $\gamma(t)$, blend $\lambda$, condition $c$
\STATE $z_0\!\sim\!\mathcal{N}(0,I)$;\; $\bar z_{-1}\!\gets\!z_0$
\FOR{$k=0$ to $K-1$}
  \STATE $\bar z_k\!\gets\!\gamma(t_k)\bar z_{k-1}+\bigl(1-\gamma(t_k)\bigr)z_k$
  \STATE $z_k^{\text{peek}}\!\gets\!(1-\lambda)z_k+\lambda\bar z_{k-1}$
  \STATE $v_k\!\gets\!v_\Theta(z_k^{\text{peek}},t_k,c)$
  \STATE $z_{k+1}\!\gets\!z_k-\eta_k v_k$
\ENDFOR
\STATE \Return $z_K$
\end{algorithmic}
\end{algorithm}

\paragraph{Reduction to Conventional Flow Sampling}
When $\lambda=0$, the blended latent in \eqref{eq:peek} reduces to the current state $z_k$, and the update rule in \eqref{eq:update} becomes identical to the standard Euler step.

\section{Complexity Analysis}
\label{sec:complexity}
Let $C_v$ denote the cost of one model (velocity field) evaluation $v_\Theta(z,t,c)$, $K$ the total number of sampling steps, and $d$ the latent dimensionality.
Both proposed schemes preserve the same asymptotic complexity as standard flow sampling, requiring only one model evaluation per step.
As summarized in Table~\ref{tab:complexity}, Look-Ahead and Look-Back introduce minor $\mathcal{O}(K,d)$ vector operations for curvature gating and exponential averaging, resulting in negligible runtime overhead and slightly higher memory usage for the running average in Look-Back.
Overall, both methods achieve improved stability and smoothness at nearly identical computational cost to the baseline.
\begin{table}[H]
\centering
\caption{Complexity comparison of different flow sampling schemes. $C_v$ denotes a single model evaluation cost; $d$ is latent dimension.}
\label{tab:complexity}
\begin{adjustbox}{width=\columnwidth}
\begin{tabular}{lccc}
\toprule
\textbf{Method} & \textbf{Model Calls / Step} & \textbf{Time Complexity} & \textbf{Extra Memory} \\
\midrule
Conventional Flow Sampling & $1$ & $\mathcal{O}(K,C_v)$ & $\mathcal{O}(d)$ \\
Look-Ahead & $1$ & $\mathcal{O}(K,C_v + K,d)$ & $\mathcal{O}(d)$ \\
Look-Back & $1$ & $\mathcal{O}(K,C_v + K,d)$ & $\mathcal{O}(2d)$ \\
\bottomrule
\end{tabular}
\end{adjustbox}
\end{table}

%% file: sec/expr.tex
\section{Experiment \& Analysis}

\begin{table*}[t]
\centering
\scriptsize
\caption{Performance of generated images produced by various methods on COCO17, CUB-200, and Flickr30K.}
\vspace{-2ex}
\label{tab:perf_comparison}
\adjustbox{width=1\textwidth}{
\begin{tabular}{llccccccc}
\toprule
\textbf{Dataset} & \textbf{Method} & \textbf{FID} {$\downarrow$} & \textbf{IS} {$\uparrow$} & \textbf{CLIPScore} {$\uparrow$} & \textbf{BLEU-4} {$\uparrow$} & \textbf{METEOR} {$\uparrow$} & \textbf{ROUGE-L} {$\uparrow$} & \textbf{CLAIR} {$\uparrow$} \\
\cmidrule(lr){1-1} \cmidrule(lr){2-2} \cmidrule(lr){3-3} \cmidrule(lr){4-4} \cmidrule(lr){5-5} \cmidrule(lr){6-9} 

\multirow{6}{*}{\textbf{COCO17} \cite{lin2014microsoft}}
 & SDv3.5 & 28.46 & 33.64 & 0.3284 & 7.99 & 29.17 & 35.11 & 71.45 \\
 & SDv3.5 w/ A-Euler \cite{jin2025flops} & 27.64 & 34.63 & 0.3284 & 7.96 & 29.09 & 35.17 & 71.23 \\
 & SDv3.5 w/ Self-Guidance \cite{li2025self} & 40.98 & 28.54 & 0.3121 & 6.52 & 26.18 & 31.59 & 62.26 \\
 & SDv3.5 w/ Momentum & 29.36 & 33.27 & \textbf{0.3329} & 8.26 & 29.43 & 35.27 & 71.41 \\
 & SDv3.5 w/ Look-Ahead & \textbf{26.17} & 34.60 & 0.3296 & \textbf{8.82} & 30.45 & 36.15 & \textbf{72.99} \\
 & SDv3.5 w/ Look-Back & 26.27 & \textbf{34.81} & 0.3294 & 8.76 & \textbf{30.53} & \textbf{36.17} & 72.77 \\ \midrule


\multirow{6}{*}{\textbf{CUB-200} \cite{welinder2010caltech}}
 & SDv3.5 & 24.92 & 4.81 & 32.40 & 0.13 & 16.57 & 17.64 & 66.69 \\
 & SDv3.5 w/ A-Euler \cite{jin2025flops} & 22.98 & 5.05 & 32.50 & 0.12 & 16.39 & 17.51 & 66.22 \\
 & SDv3.5 w/ Self-Guidance \cite{li2025self} & 61.16 & 4.98 & 31.25 & \textbf{0.18} & \textbf{17.85} & 17.78 & 62.28 \\
 & SDv3.5 w/ Momentum & 26.25 & 4.67 & 32.64 & \textbf{0.18} & 17.02 & \textbf{17.94} & 66.57 \\
 & SDv3.5 w/ Look-Ahead & 21.99 & 5.12 & 32.64 & 0.14 & 16.64 & 17.65 & \textbf{66.78} \\
 & SDv3.5 w/ Look-Back & \textbf{19.73} & \textbf{5.32} & \textbf{32.96} & 0.15 & 16.81 & 17.86 & 66.65 \\ \midrule

\multirow{6}{*}{\textbf{Flickr30K} \cite{plummer2015flickr30k}}
 & SDv3.5 & 79.58 & 17.95 & 0.3379 & 4.12 & 23.88 & 29.30 & 68.03 \\
 & SDv3.5 w/ A-Euler \cite{jin2025flops} & 78.57 & 17.83 & 0.3371 & 3.91 & 22.99 & 29.00 & 67.56 \\
 & SDv3.5 w/ Self-Guidance \cite{li2025self} & 93.07 & 14.44 & 0.3160 & 3.35 & 20.80 & 26.27 & 57.35 \\
 & SDv3.5 w/ Momentum & 79.05 & 17.56 & \textbf{0.3420} & 4.18 & 24.27 & 29.32 & 67.20 \\
 & SDv3.5 w/ Look-Ahead & 75.88 & 18.12 & 0.3388 & \textbf{4.97} & \textbf{25.32} & \textbf{30.58} & \textbf{70.16} \\
 & SDv3.5 w/ Look-Back & \textbf{75.62} & \textbf{18.30} & 0.3386 & 4.73 & 25.26 & 30.07 & 69.43 \\
\bottomrule
\end{tabular}}
\end{table*}

\newcommand{\imgrowA}[8]{%
{\small\raggedright #1} & 
\includegraphics[width=\imgw,valign=t]{\imgpath/#2_momentum.png} & 
\includegraphics[width=\imgw,valign=t]{\imgpath/#2_selfguidance.png} & 
\includegraphics[width=\imgw,valign=t]{\imgpath/#2_aflops.png} & 
\includegraphics[width=\imgw,valign=t]{\imgpath/#2_pretrained.png} & 
\includegraphics[width=\imgw,valign=t]{\imgpath/#2_lookahead.png} & 
\includegraphics[width=\imgw,valign=t]{\imgpath/#2_lookback.png} \\ 
& {\footnotesize #3} & {\footnotesize #4} & {\footnotesize #5} & {\footnotesize #6} & {\footnotesize #7} & {\footnotesize #8} \\[8pt]
}

\begin{figure*}[t]
\centering
\setlength{\tabcolsep}{2pt} 
\renewcommand{\arraystretch}{1.3} 
\begin{tabular}{>{\raggedright\arraybackslash}p{0.20\textwidth} *{6}{c}}
\toprule
\textbf{Caption} & \textbf{Momentum} & \textbf{Self Guidance} & \textbf{A-Euler} & \textbf{Pretrained} & \textbf{Look-Ahead} & \textbf{Look-Back} \\
\midrule
\imgrowA{Peace River Animals Jigsaw Puzzle}{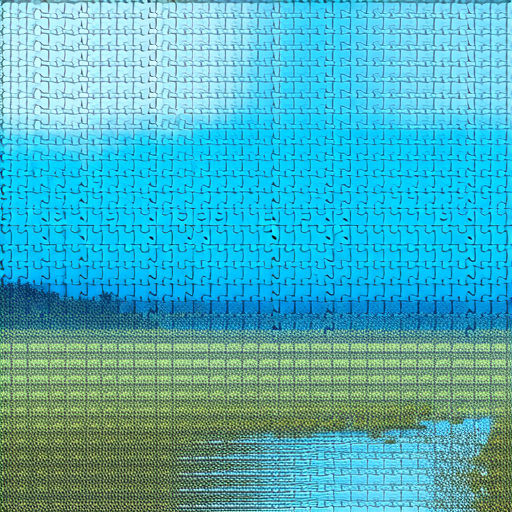}{70 \quad / \quad 0.38}{40 \quad / \quad 0.24}{40 \quad / \quad 0.34}{60 \quad / \quad 0.37}{85 \quad / \quad 0.39}{90 \quad / \quad 0.38}
\imgrowA{An intersection with a stoplight on a roadway that has no vehicles traveling on it.}{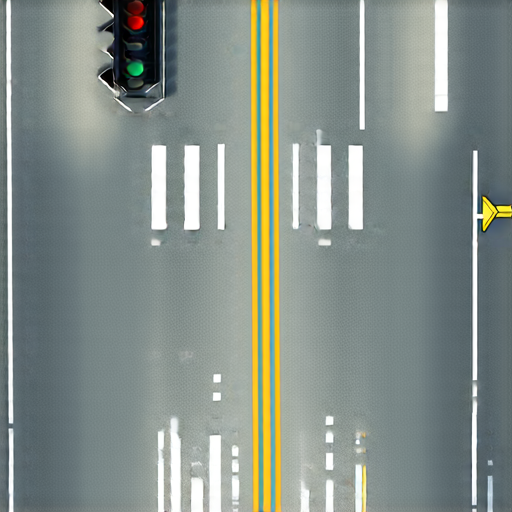}{50 \quad / \quad 0.30}{85 \quad / \quad 0.33}{30 \quad / \quad 0.28}{20 \quad / \quad 0.31}{85 \quad / \quad 0.31}{85 \quad / \quad 0.31}
\imgrowA{A person flying a kite on wet sand}{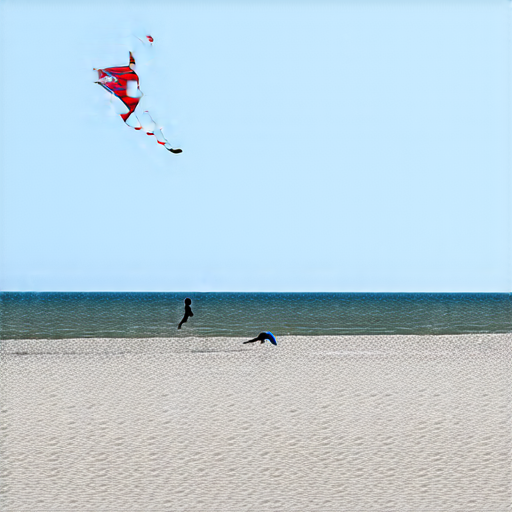}{90 \quad / \quad 0.35}{90 \quad / \quad 0.35}{90 \quad / \quad 0.34}{60 \quad / \quad 0.33}{90 \quad / \quad 0.33}{90 \quad / \quad 0.34}
\imgrowA{An green and white overhead street sign on Interstate 278 for Queens and Bronx, showing a truck restriction}{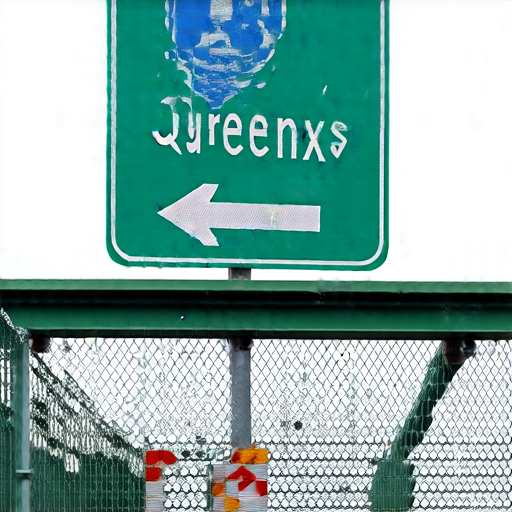}{70 \quad / \quad 0.38}{60 \quad / \quad 0.35}{60 \quad / \quad 0.41}{40 \quad / \quad 0.37}{85 \quad / \quad 0.41}{85 \quad / \quad 0.42}
\bottomrule
\end{tabular}
\caption{Qualitative comparison showing LookAhead and LookBack produce higher quality images with better coherence and detail than baseline methods. Scores shown are CLAIR / CLIPScore.}
\label{fig:qual}
\end{figure*}

\paragraph{Experimental Set-Up} 
We evaluate all methods under identical SDv3.5 inference settings, comparing the baseline sampler with the proposed Look-Ahead and Look-Back schemes, as well as prior training-free baselines A-Euler~\cite{jin2025flops} and Self-Guidance~\cite{li2025self}. All models adopt a 25-step sampling schedule with classifier-free guidance (CFG) = 7, ensuring that performance variations reflect only the stabilization behavior. The chosen configuration follows recommendations from prior studies~\cite{liu2023flow,lu2022dpm,wang2024rectified} and the HuggingFace guidelines. A single random seed is sampled once and fixed across all experiments. All hyperparameters related to the proposed methods are provided in the appendix.


We conduct evaluations on three benchmark datasets: COCO17~\cite{lin2014microsoft} (validation set), CUB-200~\cite{welinder2010caltech} (test set), and Flickr30K~\cite{plummer2015flickr30k} (test set), covering diverse image–text domains for assessing both visual fidelity and semantic alignment. All images are generated at a resolution of $512\times512$, following the protocol in \cite{ma2024sit}.

We evaluate image generation quality using FID~\cite{heusel2017fid}, IS~\cite{salimans2016improved}, and CLIPScore~\cite{hessel2021clipscore} to assess visual fidelity and semantic alignment. Additionally, captions for generated images are produced using BLIPv2~\cite{li2023blip} and compared with ground-truth captions via language-based metrics including BLEU-4~\cite{papineni2002bleu}, METEOR~\cite{banerjee2005meteor}, ROUGE-L~\cite{lin2004rouge}, and CLAIR~\cite{chan2023clair}, providing comprehensive evaluation of text–image consistency.


\paragraph{Performance}
As shown in Table \ref{tab:perf_comparison}, the proposed Look-Ahead and Look-Back schemes consistently outperform existing training-free samplers in both fidelity and semantic alignment.
On COCO17, Look-Ahead achieves the lowest FID of 26.17 and the highest BLEU-4 of 8.82, surpassing A-Euler (27.64 FID) and Momentum (29.36 FID).
Look-Back attains a similar FID (26.27) while further improving IS to 34.81 and METEOR to 30.53.
These results indicate that curvature-aware interpolation and exponential latent averaging jointly enhance the numerical stability of ODE integration, reducing oscillations without extra model evaluations.
In particular, both methods substantially outperform Self-Guidance (40.98 FID, 28.54 IS), which modifies the velocity field but often amplifies instability in rectified-flow inference.
The observed gains in CLAIR, 72.99 for Look-Ahead and 72.77 for Look-Back, demonstrate their effectiveness in maintaining high-level semantic coherence while stabilizing low-level integration dynamics.

The advantages of the proposed methods generalize across diverse datasets with distinct statistical characteristics.
On CUB-200, Look-Ahead and Look-Back reduce the FID from 24.92 (baseline) to 21.99 and 19.73, respectively. The improvement of about 5 points indicates enhanced image fidelity and better integration stability.
On Flickr30K, Look-Ahead improves BLEU-4 from 4.12 to 4.97 and ROUGE-L from 29.30 to 30.58, showing that curvature-gated correction strengthens text-image consistency.
Meanwhile, Look-Back achieves the lowest FID (75.62 vs. 79.58 baseline) and the highest IS (18.30 vs. 17.95 baseline), confirming smoother and more reliable convergence in the presence of high-curvature dynamics.
Together, these results demonstrate that the proposed Look-Ahead and Look-Back schemes not only stabilize the numerical integration but also consistently improve both visual fidelity and semantic alignment across datasets of varying complexity.

\paragraph{Qualitative Analysis}
Figure \ref{fig:qual} qualitatively highlights that the proposed Look-Ahead and Look-Back schemes produce sharper, more coherent, and semantically faithful images than all baselines.
For instance, in the Peace River Animals Jigsaw Puzzle example, both methods recover fine-grained details in the animal textures and forest background that are lost in Momentum or A-Euler, while maintaining natural composition and color consistency.
Similarly, in the Street sign example, Look-Ahead and Look-Back clearly render the ``Queens / Bronx'' text and truck icon with legible edges, whereas baselines produce fragmented signage.

\begin{figure*}[t!]
  \centering
  \includegraphics[width=0.8\textwidth]{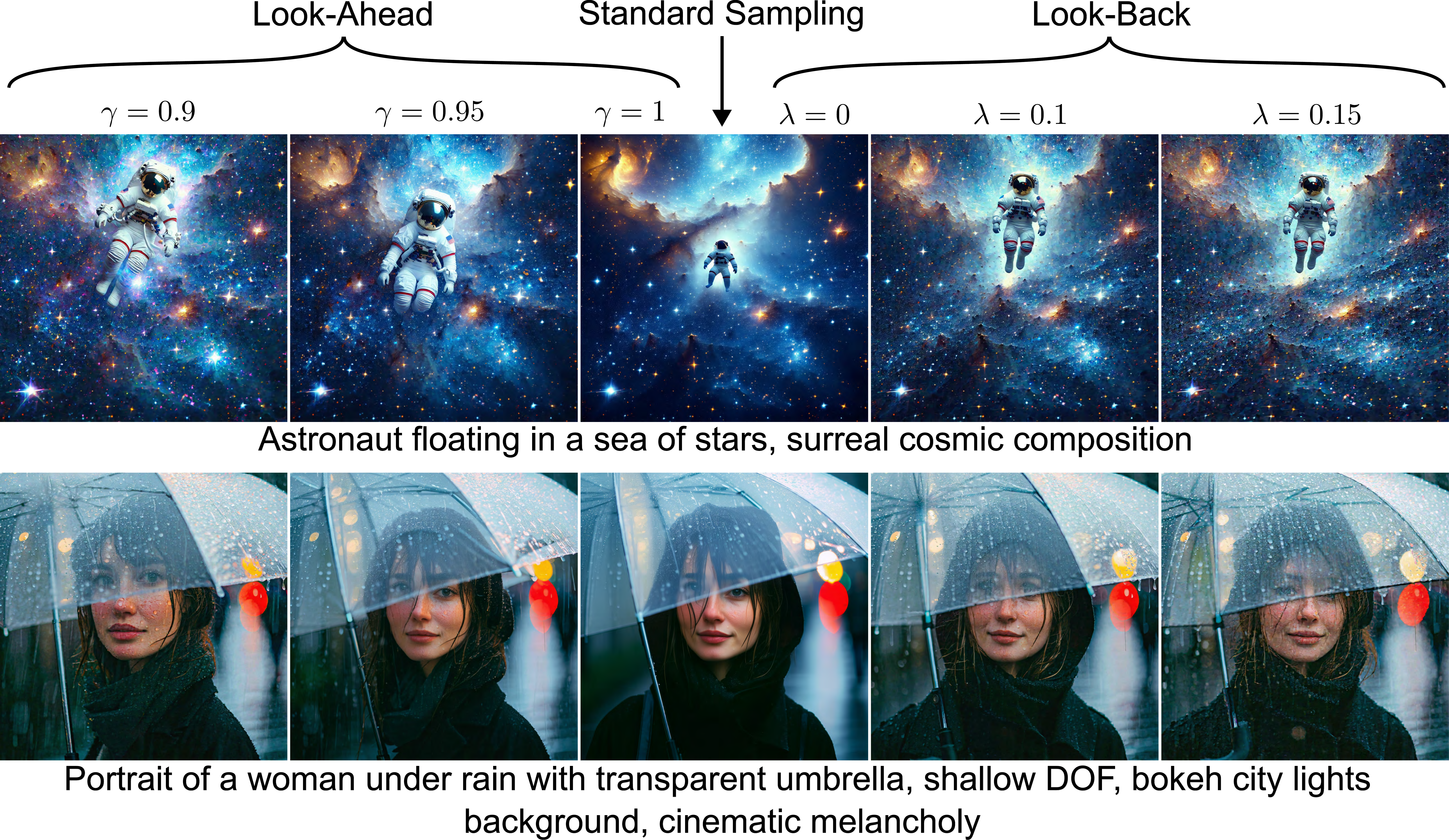}
  \caption{Visual effects of different $\gamma$ (Look-Ahead) and $\lambda$ (Look-Back). The Look-Ahead and Look-Back generations exhibit richer and more intricate visual details in the astronaut compared to the standard sampling. In the rainy portrait, the Look-Ahead and Look-Back generations produce more realistic raindrop details on the girl’s coat, making the scene more consistent with a rainy atmosphere, whereas the standard sampling fails to capture such effects. 
  }
  \label{fig:effect_proposed}
\end{figure*}

Figure \ref{fig:effect_proposed} reveals clear qualitative advantages of the proposed schemes over standard sampling.
For the astronaut composition, Look-Ahead and Look-Back generate markedly richer cosmic scenes with more intricate starfield details and refined suit textures, whereas standard sampling produces a flatter, less detailed result. In the portrait, the key difference lies in the realistic raindrop details rendered on the woman's coat by both proposed methods, an atmospheric nuance completely absent in the standard sampling.

\begin{figure}[t]
    \centering
    \setlength{\tabcolsep}{2pt}
    \begin{tabular}{cc}
        \begin{subfigure}[b]{0.48\columnwidth}
            \centering
            \includegraphics[width=\linewidth]{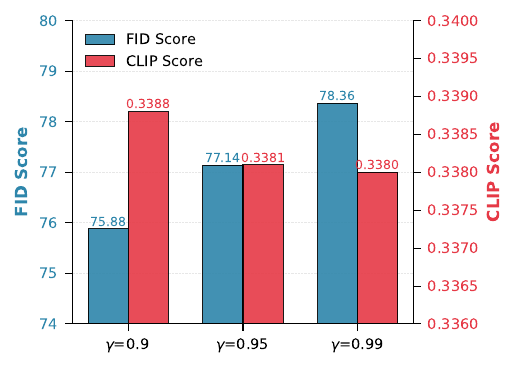}
            \caption{Look-Ahead ($\tau_{\text{curv}}=1$).}
            \label{fig:subfig1}
        \end{subfigure} &
        \begin{subfigure}[b]{0.48\columnwidth}
            \centering
            \includegraphics[width=\linewidth]{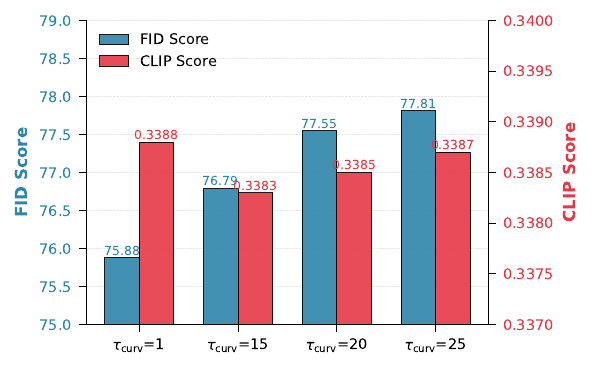}
            \caption{Look-Ahead ($\gamma=0.9$).}
            \label{fig:subfig2}
        \end{subfigure} \\[4pt]
        \begin{subfigure}[b]{0.48\columnwidth}
            \centering
            \includegraphics[width=\linewidth]{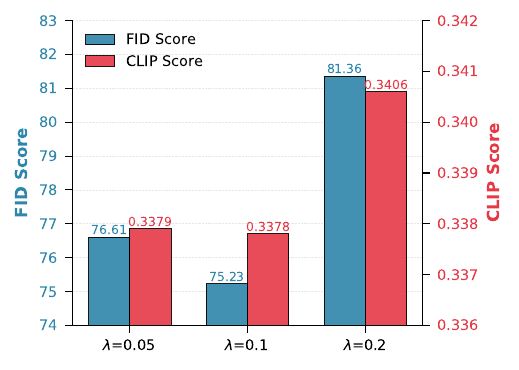}
            \caption{Look-Back ($\xi_{\ast}=0$).}
            \label{fig:subfig3}
        \end{subfigure} &
        \begin{subfigure}[b]{0.48\columnwidth}
            \centering
            \includegraphics[width=\linewidth]{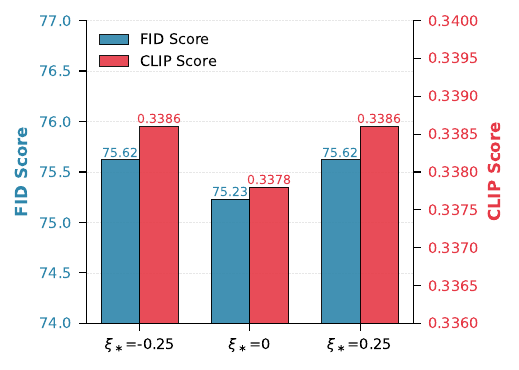}
            \caption{Look-Back ($\lambda=0.1$).}
            \label{fig:subfig4}
        \end{subfigure}
    \end{tabular}
    \caption{Ablation study of key hyperparameters related to the proposed methods on Flickr30K.}
    \label{fig:ablation}
\end{figure}

\noindent\textbf{Ablation Study} Figure \ref{fig:ablation} analyzes the sensitivity of the proposed Look-Ahead and Look-Back schemes to their key hyperparameters, revealing how curvature gating and exponential averaging jointly govern stability and fidelity during ODE integration. In the Look-Ahead scheme, both the curvature threshold $\tau_{\text{curv}}$ and interpolation factor $\gamma$ control the balance between numerical conservativeness and trajectory progress. As shown in Figures \ref{fig:ablation}(a)–(b), excessively large $\tau_{\text{curv}}$ values permit unstable full-step updates resembling Euler behavior, while overly restrictive thresholds hinder convergence; the optimal regime emerges around $\tau_{\text{curv}}=1$ and $\gamma=0.9$, which achieve the lowest FID (75.88) and highest CLIPScore (0.3388), confirming that moderate curvature moderation yields the best trade-off between efficiency and smoothness. In the Look-Back scheme, Figures \ref{fig:ablation}(c)–(d) illustrate that small blending weights and a centered SNR midpoint ($\xi^*=0$) produce the most stable results. Larger $\lambda$ over-smooths latent trajectories and biases the generative path, whereas proper SNR-dependent decay $\gamma(t)$ ensures that smoothing fades near convergence, allowing high-frequency details to re-emerge. Collectively, these ablations demonstrate that both schemes are robust within a broad parameter range and validate the importance of curvature-aware moderation and adaptive temporal averaging for stable training-free flow sampling.

\paragraph{Runtime}
Under identical settings, SDv3.5 requires 11.0 seconds per image, while Look-Ahead and Look-Back take 11.4 and 11.2 seconds, respectively, introducing only a marginal overhead.

%% file: sec/conclusion.tex
\section{Conclusion}


We presented two complementary training-free latent-trajectory smoothing methods, Look-Ahead and Look-Back, which refine the flow-matching generative process directly in latent space. Unlike prior approaches that modify the velocity field and risk accumulating errors along the sampling path, our latent-trajectory adjustments benefit from the intrinsic corrective behavior of the pretrained velocity network, leading to more stable and robust generation. Look-Ahead leverages curvature-gated averaging with future latents, while Look-Back applies an exponentially decayed moving average informed by past states. Extensive experiments on COCO17, CUB-200, and Flickr30K show that our methods consistently outperform strong state-of-the-art baselines across a wide range of metrics. These results highlight latent-trajectory smoothing as an effective and general strategy for improving training-free flow-based image generation.

%% file: sec/appendix.tex
\section{Sampling with Momentum}

For each sample trajectory, we maintain a first-moment vector $m_k$ initialized to zero.
At each step, we update this momentum using an exponential moving average:
\begin{align}
g_k &= -\,v_\Theta(z_k, t_k, c), \\
m_{k+1} &= \beta_1\,m_k + (1-\beta_1)\,g_k, \\
z_{k+1} &= z_k + \eta_k\,m_{k+1},
\label{eq:momentum_update}
\end{align}
where $\beta_1\in[0,1)$ controls the momentum strength (e.g., $0.8$ by default).
This formulation introduces temporal smoothness in velocity directions and stabilizes
sampling for stiff or high-curvature trajectories.


This training-free method maintains momentum independently for each trajectory without batch sharing, introducing negligible computational overhead by requiring only one additional vector state per sample. Notably, in the specific case where $\beta_1=0$, the algorithm reduces exactly to the vanilla Euler solver. Empirically, we observe that $\beta_1=0.8$ achieves the optimal performance on Flickr30K.

\section{Pseudo Algorithm}

\begin{algorithm}[H]
\caption{Sampling with Momentum}
\label{alg:momentum_sampling}
\begin{algorithmic}[1]
\REQUIRE Trained velocity field $v_\Theta$, condition $c$, time grid $1=t_0>\cdots>t_K=0$, step sizes $\eta_k$, momentum coefficient $\beta_1$
\STATE Sample $z_0 \sim p_1$ \COMMENT{initialize from noise prior}
\STATE $m_0 \leftarrow 0$
\FOR{$k = 0, 1, \dots, K-1$}
    \STATE $g_k \leftarrow -v_\Theta(z_k, t_k, c)$
    \STATE $m_{k+1} \leftarrow \beta_1 m_k + (1-\beta_1) g_k$
    \STATE $z_{k+1} \leftarrow z_k + \eta_k m_{k+1}$
\ENDFOR
\STATE \Return $z_K$ \COMMENT{final generated sample}
\end{algorithmic}
\end{algorithm}

Fundamentally, this approach operates as a form of trajectory smoothing.
The method acts as a temporal low-pass filter on the flow velocity, suppressing
abrupt direction changes (high-frequency oscillations) while preserving the global
flow geometry. By aggregating historical directional information, the momentum term
effectively straightens the integration path, making it robust to local irregularities
or noise in the learned velocity field. This avoids the need to assume diminishing
gradients and ensures the solver smoothly reduces to Euler when $\beta_1 \to 0$,
providing convergence stability with minimal overhead.

\section{Complexity Analysis}
\label{sec:complexity}

We now analyze and compare the computational complexity of three flow sampling schemes, i.e., conventional, Look-Ahead, and Look-Back, in terms of their dominant operations per timestep. Let $C_v$ denote the cost of one model (velocity field) evaluation $v_\Theta(z,t,c)$, and $K$ the total number of sampling steps.

\paragraph{Conventional Flow Sampling}
Each step directly evaluates the velocity field once and performs a single latent update
Eq. (5).
Hence, the total cost scales linearly with $K$:
\[
\mathcal{C}_{\text{flow}} = K\,C_v.
\]

\paragraph{Look-Ahead Sampling}
The proposed Look-Ahead scheme estimates the peek velocity $\tilde v$ via a finite-difference extrapolation Eq. (6) without invoking an additional model call.
Thus, its computational complexity remains identical to the conventional solver, but with extra $\mathcal{O}(d)$ vector arithmetic per step (where $d$ is latent dimensionality), which is negligible compared to $C_v$.
\[
\mathcal{C}_{\text{LA}} = K\,C_v + \mathcal{O}(K\,d).
\]
In practice, the runtime overhead is marginal ($<3\%$), as the dominant cost remains model evaluation.

\paragraph{Look-Back Sampling}
The Look-Back scheme introduces exponentially weighted averaging and blending operations Eq. (9)–Eq. (10), both linear in $d$.
Like Look-Ahead, it performs one velocity-field query per step; thus its asymptotic complexity also matches the baseline:
\[
\mathcal{C}_{\text{LB}} = K\,C_v + \mathcal{O}(K\,d).
\]
However, its memory footprint is slightly higher due to storage of the running average $\bar z_k$.

Table~\ref{tab:complexity} summarizes the comparative complexity, model calls, and memory cost.
Both Look-Ahead and Look-Back retain the same asymptotic complexity as conventional flow sampling, offering stability or smoothness improvements at nearly constant computational cost.

\begin{table}[H]
\centering
\caption{Complexity comparison of different flow sampling schemes. $C_v$ denotes a single model evaluation cost; $d$ is latent dimension.}
\label{tab:complexity}
\begin{adjustbox}{width=.8\columnwidth}
\begin{tabular}{lccc}
\toprule
\textbf{Method} & \textbf{Model Calls / Step} & \textbf{Time Complexity} & \textbf{Extra Memory} \\
\midrule
Conventional Flow Sampling & $1$ & $\mathcal{O}(K\,C_v)$ & $\mathcal{O}(d)$ \\
Look-Ahead (Curvature-Gated) & $1$ & $\mathcal{O}(K\,C_v + K\,d)$ & $\mathcal{O}(d)$ \\
Look-Back (EMA-Stabilized) & $1$ & $\mathcal{O}(K\,C_v + K\,d)$ & $\mathcal{O}(2d)$ \\
\bottomrule
\end{tabular}
\end{adjustbox}
\end{table}

The additional $\mathcal{O}(K\,d)$ arithmetic in Look-Ahead and Look-Back schemes is computationally negligible in practice, since $d$ corresponds to the latent dimensionality of the VAE or diffusion backbone rather than the pixel space. Typical latent sizes (e.g., $d\!\approx\!4\times64\times64$ for Stable Diffusion) are orders of magnitude smaller than the full image dimensionality. Consequently, the total overhead $K\,d$ remains well within modern GPU memory and compute budgets even for large $K$ (e.g., $K\!\le\!100$). This ensures that both Look-Ahead and Look-Back sampling remain fully feasible and efficient in high-resolution generative pipelines, maintaining near-identical runtime to the baseline solver while providing enhanced stability and smoothness.



\section{Implementation Details}

CLAIR~\cite{chan2023clair} leverages a Large Language Model to evaluate semantic equivalence. For the CLAIR implementation, we utilize the OpenAI gpt-4.1-mini model as the evaluator backend.

To ensure reproducibility, we detail the specific hyperparameter configurations used for the Look-Ahead and Look-Back schemes across the evaluated datasets (COCO17, CUB-200, and Flickr30K). For the Look-Ahead scheme, the curvature threshold $\tau_{\mathrm{curv}}$ and the interpolation factor $\gamma$ are tuned to balance trajectory fidelity with sampling efficiency; these settings are provided in Table~\ref{tab:hyper_lookahead}. For the Look-Back scheme, we specify the blending weight $\lambda$ and the SNR midpoint $\xi_\ast$, which jointly control the intensity of the history-based stabilization and the decay scheduling. These values are listed in Table~\ref{tab:hyper_lookback}. Regarding the dataset protocols, since both COCO17 and Flickr30K provide five captions per image, we use the specific prompt that achieves the best CLIPScore for evaluation purposes.

\begin{table}[t]
\centering
\scriptsize
\caption{Key hyperparameters related to the proposed look-ahead scheme.}
\label{tab:hyper_lookahead}
\adjustbox{width=.2\textwidth}{
\begin{tabular}{lccc}
\toprule
\textbf{Dataset} & \textbf{$\gamma$} & \textbf{$\tau_{curv}$}  \\
\cmidrule(lr){1-1} \cmidrule(lr){2-2} \cmidrule(lr){3-3}
COCO-2017 & 0.9 & 10 \\
CUB-200 & 0.95 & 1 \\
Flickr30K & 0.9 & 1 \\
\bottomrule
\end{tabular}}
\end{table}

\begin{table}[t]
\centering
\scriptsize
\caption{key hyperparameters related to the proposed look-back scheme.}
\label{tab:hyper_lookback}
\adjustbox{width=.2\textwidth}{
\begin{tabular}{lcc}
\toprule
\textbf{Dataset} & \textbf{$\lambda$} & \textbf{$\xi_\ast$}  \\
\cmidrule(lr){1-1} \cmidrule(lr){2-2} \cmidrule(lr){3-3} 
COCO-2017 & 0.1 & 0 \\
CUB-200 & 0.1 & 0 \\
Flickr30K & 0.1 & 0.25 \\
\bottomrule
\end{tabular}}
\end{table}